# Towards System Modelling to Support Diseases Data Extraction from the Electronic Health Records for Physicians' Research Activities

Bushra F. Alsaqer[1] , Alaa F. Alsaqer[1] , and Amna Asif [2]

1. Department of Information Systems,
College of Computer Science and Information Technology,
King Faisal University, Al-Hufof, Al-Hasa, Saudi Arabia
2. Lancaster University Leipzig, Germany
bushra.f.s@hotmail.com, afalsaqer@windowslive.com, and a.asif2@lancaster.ac.uk

*Abstract*

*The use of Electronic Health Records (EHRs) has increased dramatically in the past 15 years, as, it is considered an important source of managing data od patients. The EHRs are primary sources of disease diagnosis and demographic data of patients worldwide. Therefore, the data can be utilized for secondary tasks such as research. This paper aims to make such data usable for research activities such as monitoring disease statistics for a specific population. As a result, the researchers can detect the disease causes for the behavior and lifestyle of the target group. One of the limitations of EHRs systems is that the data is not available in the standard format but in various forms. Therefore, it is required to first convert the names of the diseases and demographics data into one standardized form to make it usable for research activities. There is a large amount of EHRs available, and solving the standardizing issues requires some optimized techniques. We used a first-hand EHR dataset extracted from EHR systems. Our application uploads the dataset from the EHRs and converts it to the ICD-10 coding system to solve the standardization problem. So, we first apply the steps of pre-processing, annotation, and transforming the data to convert it into the standard form. The data pre-processing is applied to normalize demographic formats. In the annotation step, a machine learning model is used to recognize the diseases from the text. Furthermore, the transforming step converts the disease name to the ICD-10 coding format. The model was evaluated manually by comparing its performance in terms of disease recognition with an available dictionary-based system (MetaMap). The accuracy of the proposed machine learning model is 81%, that outperformed MetaMap accuracy of 67%. This paper contributed to system modelling for EHR data extraction to support research activities.*

## I. INTRODUCTION

There are various and different types of medical information sources, namely, Hospital Information Systems (HIS), Electronic Medical Records (EMR), or Electronic Health Records (EHRs). HIS is a system that manages medical, administrative, financial and legal features of a hospital. EMR grant the possibility to store all relevant patient information in electronic format. This information may hold symptoms, notes, remarks made by one or more physicians, and descriptions of relevant patient events [1]. EHRs are an electronic repository of patient data which was designed based on an international standard to be used by authorized users in hospitals, such as nurses and physicians. Therefore, the EHRs offer a window into the medical care, status, and outcomes of a diverse population that is representative of actual patients. The secondary use of medical data collected in EHRs is a promising step towards increasing patient-centred research and speeding the rate of new medical discoveries [2]. Accordingly, the EHRs are a valuable resource for clinical studies and research activities [3] and [4]. Physician researchers seek to develop medical sciences through research studies on the causes and prevalence of diseases using medical data extracted from the EHRs.

Yet diseases data are one of the most essential medical elements of the EHRs, since being used widely for research. Diseases and demographic data of patients such as age and gender are useful for monitoring diseases factors and estimating future diseases prevalence rates. In contrast, the multiplicity of local diseases classifications and using various medical data formats from one medical source to another make it difficult to use the EHRs data for patient history, disease causes and prevalence research [5]. So, toward making the EHRs data usable for research, there is an advanced and standard classification system utilizes disease data effectiveness for the secondary use which is International Classification of Diseases (ICD). ICD is a medical classification developed by the World Health Organization (WHO) to be used as standard diseases classification. The ICD allocates a specific symbol (code) for each disease, to simplify diseases inventory, statistics, and storage [6]. The ICD makes it easier to handle standard diseases data rather than using random and multiple diseases classifications using various disease formats.

For example, writing 'Stomach Cancer ' or 'Gastric Cancer ' to indicate stomach neoplasms, leading using various formats for the same disease. The ICD will allocate a standard name and code for this disease which is 'C16 - Malignant neoplasm of stomach'. ICD Codes are organized hierarchically with top and bottom levels. The top-level entries are general groupings, for example, "Diseases of the circulatory system", whereas the bottom level codes indicate specific symptoms or diseases and their location like "Chronic rheumatic heart diseases". Therefore, ICD provides an effective way to unify diseases data between different sources as an international classification and provides an accurate specification of the disease category. Therefore, people tend to support the EHRs by using international standard coding for diseases classification [7]. However, manual ICD coding increases the costs of doctors' training and consume more time to search for the desired code among thousands of codes, which may expose the doctor to choose the wrong code. Only 60% ~ 80% of the assigned ICD codes reflect the exact patient medical diagnosis [8].

The remaining content of this research is organized as follows: In Section II, the Related literature is presented. In Section III, the Methodology for the system is given. Section IV presents the Implementation for the system. In Section I, the Results and Evaluation for the system are given. This is followed by a Conclusion in Section V and Future Work in Section VI.

## II. RELATED RESEARCH

### A. Using Medical Data for Secondary Use

Recent studies show that EHRs have been widely utilized for medical and research activities [3] and [9]. However, there are major issues found in row EHRs related to data quality including the variety of recording EHRs data, which lead to randomness as well as misleading problems [3]. Due to this issue, other studies emphasis that EHRs require improvement, particularly in terms of data exchange to support such the secondary use purposes [9]. Since using medical data for secondary use is one of the most critical issues, it has been discussed widely in many recent studies. The studies found that demographics and diagnosis are the most captured medical elements for research, but they need to be processed and standardized to be suitable for this purpose [4] [10]. Therefore, the studies aimed at identifying problems in demographics and diagnosis data seeking for developing some systems to solve these problems. Regarding diagnosis data, diagnoses are required to be in a standardized format to facilitate diagnosis statistics and other research activities such as knowledge graphs and disease-related models. ICD is considered a solution for this problem as it is an authoritative classification system of various diagnosis and conditions and it provides more accurate diseases classification and retrieval of medical records.

Considering the complicated and dedicated manual ICD coding, some studies have proposed methodologies of automatic ICD coding. A study developed a hierarchical deep learning model with attention mechanism which can automatically select and assign ICD-9 codes given free-text diagnosis description. The experimental results showed good performance, but it could be improved further considering the noisy format of the electrical discharge summary. This limitation could be overcome by more elaborate pre-processing of diagnosis extraction and cleaner corpus with high-quality diagnosis descriptions in addition to using a newer version of ICD instead of ICD-9 [11]. In this regard, a study has proposed a methodology of deriving a semantic index for clinical text such as clinical eligibility documents based on a controlled vocabulary of frequent tags. The frequent tags are automatically mined from the text using trained tagging model and a medical source of terms that most relevant to the clinical trial domain using n-grams (i.e., continuous sub-sequences of n-words) algorithm. Each n-gram is matched against the term's lexicon. For example, the text of ''malignancy within the past 5 years'' is considered a valid n-gram because at least one word ''malignancy'' that presented in the part of the term's lexicon will be considered, even if the entire sentence is not. However, that study restricts the tagging system with lexicon terms that relevant to the clinical trial, which makes disease extraction dependent on only specific medical terms [12].

Beside diagnosis data, demographics include patients' diagnosis date, gender, age, and more, most of them can be represented in multiple formats in the EHRs making them need to be identified and normalized into a standard format [13]. A study has stated that demographics identification is crucial and indispensable for a better understanding of patients' problems such as the patient's age. Age is a key dimension for dividing the population group which can be represented in three patterns including age number, e.g. 60yrs old; age range, e.g. 40-49 years; and age term, e.g. child. For each age pattern, the study applied building corresponding rule sets approach to extract these patterns and set a standard format for them. For example, the age number identification needs a regular expression of the digits or time related words, i.e. year(s), yr(s), month(s) to build a standard number pattern. This study provided a good rule-based approach to extract and normalize age patterns in a flexible way that can be applied to other demographics elements in the EHRs such as gender or diagnosis date [14].

In general, extracting diagnosis and demographics data from medical systems for research requires a methodology that can be included in a unified system to support data processing and standardization such as (MEDSEC) system for identifying Protect Heath Information (PHI). MEDical records for SECondary use (MEDSEC) system are a proposed system that converts paper-based health records into de-personalised and de-identified documents automatically which can be accessed by secondary users without compromising the patients' privacy. MEDSEC lies on four phases, it converts images of health documents into machine-readable strings to annotate PHI using annotation rules. After that, it transforms the annotated PHI components into a standard format of Health Level 7 Clinical Document Architecture (HL7 CDA) using matching conditions. The last phase is de-identifying the records to make them available for privacy-preserving secondary use. MEDSEC shows good results but leaves an area for enhancements due to OCR and PHI annotation module errors [4]. However, MEDSEC provides an integrated model for this research methodology, including pre-process age, gender, and diagnosis date, and then annotates disease names from the diagnosis text. After that, it will transform the recognized diseases into ICD forms automatically. Then, the annotation model will be evaluated to measure the accuracy of the performance. Finally, the output of standard diagnosis data and demographics will be presented as statistics in an interactive visualization platform to support research activities. The next sections will discuss techniques and models that can be applied to this paper in detail.

*B. Data Pre-processing*

Regarding EHRs data, measuring their suitability for research can be applied by measuring medical data quality dimensions and determining what may affect these dimensions to make this data unsuitable for research use. A study analysed 95 articles to understand how to determine the appropriateness of the EHRs data for research based on data quality and types of data. Depending on the results shown in the study, completeness and accuracy are the two most important dimensions of the EHRs data quality that are usable for research [2].

There are different data pre-processing techniques, such as data cleaning, which are used to remove noise and correct inconsistencies in data. Data integration makes data that is extracted from multiple sources merged into a coherent data store such as a data warehouse. Data reduction is responsible for reducing data size by, for instance, aggregating, eliminating redundant features, or clustering. Data transformations such as normalization are used to scale data to fall within a smaller range. Accordingly, EHRs data such as demographics and diagnosis data must be cleaned to overcome their completeness challenges by removing missing values and normalized to unify their various formats to improve their accuracy. For example, data cleaning may involve transformations to correct inappropriate data, such as by transforming all entries for such fields to a standard format. Transforming medical data to a standard format depends on the form of data available. For demographic data, gender data may have been recorded using various formats. Where female gender could be expressed as ''F'', ''women'', or ''female'' depending on their system design, so they need to be normalized into a standard pattern like "Female" using the rule-based regular expression. Therefore, data pre-processing techniques can work together to improve algorithms and increase EHR data accuracy and efficiency [15] [12]. Besides that, diagnosis in the EHRs is written in free text fields, which will require more complex processing to recognize diseases names in the diagnosis text to be annotated as diseases.

*C. Diagnosis Names Recognition*

The recognition of named entities or concepts, such as diagnoses, medical problems, tests, and treatments, is a process called Named Entity Recognition (NER). NER is a basic initial step in information extraction from clinical records. Diagnosis is often described in medical records as free text using some phrases, making disease recognition (which practically means assigning standardized classification codes to diseases' names) a crucial natural language processing (NLP) challenge. Many different methods of NER in the medical field have been developed, but no single method has yet been shown to perform best generally. There are some annotation systems that reflect a variety of approaches to the recognition of named entities; two of them are UMLS Structures (Dictionary-based systems) and Statistical-based (Rule-based) systems [16] [17].

UMLS Structures are programs that recognize Unified Medical Language System (UMLS) concepts in medical text, such as MetaMap. MetaMap is a dictionary-based system that was developed to identify concepts from the UMLS in biomedical text. MetaMap depends on a minimal commitment parser texts to split them into chunks, in which UMLS concepts are identified. For statistical-based systems, they are programs that are trained to annotate medical problems, tests, and treatments in clinical records using rule-based components such as MEDLEE [16]. MedLEE is a program that was developed to extract, structure and encode clinical information from textual patient reports [18]. There is a challenge with the previously described medical NLP tools, which is that they are not easy to adapt, generalize and reuse by an unrelated institution. The reason behind that is that medical NLP programs are often tailored to domain or institution-specific document formats and other text characteristics. In addition, the intellectual property of NLP software, of course, has also been an obstacle to sharing for different platforms [19].

Consequently, to determine the most suitable system (dictionary-based or statistical-based NLP systems) to be used as a tool for our paper, both systems must be compared based on two important criteria: availability and the possibility of re-training. Regarding our paper objectives, there are two major points that must be taken into consideration during the comparison part of both systems. First, our paper is concerned with extracting diseases terms from specific diagnosis descriptive patterns in the EHRs, which differ from descriptive patterns in medical documents and reports. Second, our paper is based on making the process of disease extraction as one of the phases of an integrated system. Thus, the disease recognition system (dictionary or statistical-based system) should be used as a tool embedded in our system to train on the EHRs' diagnosis descriptive patterns and not as an independent program. Table 1 shows a comparison between dictionary-based systems and statistical-based systems in terms of availability and the possibility of re-training a system:

TABLE1: NLP SYSTEMS.

| NLP systems | Availability | Training |
|---|---|---|
| Dictionary-based systems | These systems are provided only under a license construction making them unsuitable for real time processing. | These systems cannot be automatically trained for a particular task. |

| NLP systems | Availability | Training |
|---|---|---|
| Statistical-based systems | The systems are not publicly available due to intellectual property of the software. | Although these systems generally perform very well, many contain rule-based components that cannot be easily trained and may require considerable efforts to adjust it for the task at hand. |

As noted in the comparison, both dictionary and statistical based systems can't be used for our paper, so it will use a machine learning approach for diagnosis names recognition task. Machine learning has the advantage of identifying a model of unstructured data, such as data in free text form. Also, deep learning has produced extremely promising results for many tasks in natural language understanding [20]. Therefore, this paper will retrain a model for extracting data terms from text to develop a model of diseases names recognition. The model will be able to recognize diseases from descriptive patterns using a sample of diagnosis descriptive text extracted from the EHRs to facilitate transforming diseases into a standardized classification automatically.

*D. Transforming Diseases to Standard Form*

Some studies proposed an automated ICD based coding help system using UMLS dictionary-based indexing system [21]. UMLS clusters many medically controlled vocabularies and classifications, such as medical subject headings (MeSH), to provide links between several vocabularies. An automated ICD-10 system was proposed to apply a MeSH-based indexing system and a mapping between MeSH and ICD-10 extracted from the UMLS. The indexing system can find and extract each textual element referring to MeSH terms in a whole medical record. Then, it allocates a score depending on the text length and the frequency of each term to each MeSH term extracted, which is called the indexing score. The mapping process between MeSH and ICD-10 produces a list of ICD-10 terms from a list of MeSH terms that are supposed to be equivalent. The goal of the ICD-10 classification is to make up the list of all the diagnoses. MeSH/ICD-10 mapping has been evaluated to be limited to 8% of the ICD-10 classification codes in contrast with physicians who can code with the whole classification. The limitation of MeSH/ICD-10 is due to the differences in structuration and the goal of the classifications that provide a loss of information in the mapping. Furthermore, the wording of the medical reports can affect the mapping results [22].

Hence, to avoid these limitations, the ICD mapping process should be conducted as NEL after NER (machine learning) of diseases terms. So that only the extracted recognized disease is linked with the knowledge base of ICD rather than the whole report or document. The ICD mapping process will be conducted between the data (recognized diagnosis names) and ICD knowledge base directly using an application program interface API of ICD without the need for UMLS based indexing. NEL is a task of linking named entities (recognized disease terms) into a knowledge base (ICD), which is a list of named entities and information about them. Where ICD API allows programmatic access to ICD data with the feature of ranking the codes depending on the most opportunity result of diagnosis term [23] [24].

*E. Visualization of Standard EHRs Data*

Getting knowledge from visualized data is an important area, especially for research purposes. Thus, using visualization techniques involves understanding the data that will be presented and its size, as well as identifying the target audience who will extract the data. That is because visualizing data carries the information in the best and easiest way to the audience or end users [25]. The diagnosis data involves various diagnoses such as kidney, heart, diabetes, and tumours, which are closely related to demographics. For gender, medicine has shown that there are diagnoses that women are more likely to have than men, such as 'Hyperthyroidism'. In terms of age, medicine has also shown that 'Insulin-dependent diabetes mellitus' affects children more than adults. Diagnosis date is also a key dimension, which is considered as a measure of the incidence of some diseases incrementally. These data are an important source for the researcher if they are standardized, annotated and presented on a platform through visualization techniques.

Since this paper focuses on providing the EHR data to support research and analysis studies, data analysis is essential for selecting appropriate visualization techniques. Analysis techniques must be satisfactory and adequate for the type of data, which is disease statistics, and require an interactive visualization technique to provide users with real-time responses to extract more knowledge and easier information analysis. Interactive visualization involves examining how humans interact with computers to create graphic representations of knowledge and how this process can be made more useful and efficient.

Accordingly, our research needs to satisfy two criteria for visualizing diseases statistics considering demographics to be interactive. First, there needs to be human input: control of some features of the visual representation of data or information must be available to a human. For example, the disease information and medical statistics must be presented to the researcher or physicians to extract knowledge about disease percentages from one year to another and for males and females. These options must be provided for users to make the presentative data interactive with them. They will be able to select disease information according to some selected factors. Secondly, there should be a response time for any changes the user makes. For example, if the user selects a specific date to seek certain disease statistics, the information will be presented dependently [26]. This paper will meet the need for combining EHR data in ICD form with an interactive visualization platform to support research activities.

III. METHODOLOGY

This section presents the methodology used and the proposed model of extracting, standardizing, and visualizing the diagnosis and demographics data from the EHRs. Thus, the paper seeks to develop a system that applies some Natural Language Processing (NLP) techniques to the EHR data, including normalization, annotation, and transformation. As shown in Fig.1, the proposed model is based on five phases, including extracting raw data from the EHRs, pre-processing the data, annotating the diseases, transforming the diseases to ICD-10, evaluating the accuracy, and presenting the standard data in the interactive dashboard.

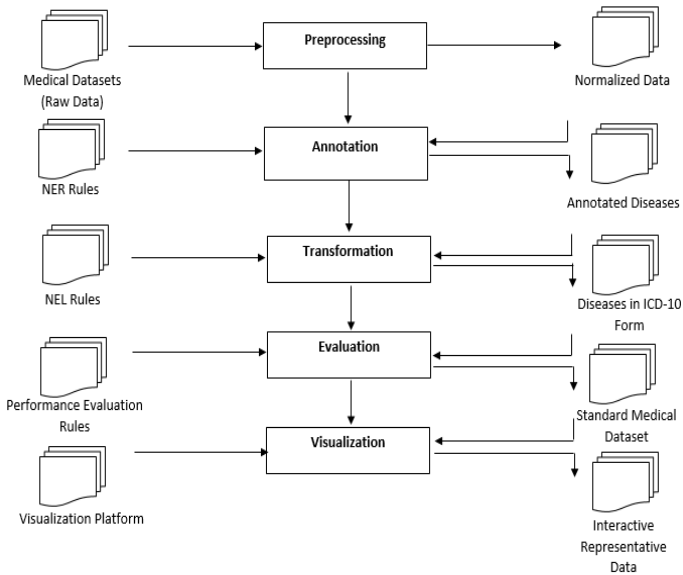

Fig. 1. Proposed Model

Pre-processing phase applies normalization processes on the raw data which includes demographics and diagnosis data. Normalizing data applied to make the data normalized, complete and correct. Annotation process is responsible for determining disease name in the given diagnosis text using machine learning recognition approach called Named Entity Recognition (NER). Recognizing diseases names from the text is facilitate transforming them into standard ICD-10 form. Transformation phase is responsible for taking recognized diseases names and converting them to ICD-10 form. Transformation is using Named Entity Linking (NEL) approach and ICD knowledge base (KB) to find ICD-10 codes with its name for each recognized disease. The output of this phase is standard diagnosis data. The evaluation phase will compare the accuracy of the diseases recognition result of the machine learning NER and MetaMap (dictionary-based) recognizer. The visualization phase is responsible for presenting the standard medical dataset in an interactive visualization platform as standard diseases statistics for research activities.

A. Data Component

Two datasets, the medical dataset and the diagnosis dataset, will be used for system development. The medical dataset will be used to develop the system, and the diagnosis dataset will be used to generate a machine learning model for disease recognition.

The medical dataset includes a sample of 300 patients in an Excel file that has been collected from multiple EHRs of medical polyclinics. One of the limitations of existing EHRs from polyclinics is that it does not allow automatic extraction of medical data from EHRs records due to security issues. Therefore, the medical data was collected manually with the permission of the clinic head. The medical polyclinics offers multiple clinics such as General clinic, Paediatric clinic, and Internal clinic, to extract various diagnosis statements and different demographics. The medical dataset included four attributes: Age, Gender, Diagnosis Date, and Diagnosis as shown in Table 2.

TABLE 2: A SAMPLE OF MEDICAL DATASET.

| Gender | Age | Diagnosis | Diagnosis Date |
|--------|-----|-----------|----------------|
| F | 20 | Cystitis | 9/4/1439 |
| Female | 56 | The result of the analysis is Sickle-Cell Anaemia | 10/4/1439 |
| F | 21 | Gastroenteritis | 10/4/1439 |
| F | 34 | Stage 5 of Chronic Kidney Disease | 10/4/1439 |
| f | 18 | Cystitis | 10/4/1439 |
| female | 40 | Ischemic Heart Disease | 5 years ago |
| M | 60 yrs | New discovered hypertension + stroke | 8/5/1439 |
| F | 22 yr | uncontrol Insulin-dependent diabetes mellitus | 8/5/1439 |
| M | 6 yr | Autism | 29/1/1439 |
| F | 21 | Follow up Diabetes mellitus type I /Primary hypothyroidism | 8/5/1439 |

As shown in Table 2 demographics values are expressed in various formats due to using a predefined list or free text fields. For example, age is expressed as "20" or "60 yrs" and gender is expressed as "f" or "female". Also, diagnosis date is expressed in date or words formats. Therefore, demographics need to be normalized into standard formats to be used for diseases statistics. While diagnosis is expressed in free text including disease name only such as " Ischemic Heart Disease ", or diagnosis description such as " The result of the analysis is Sickle-Cell Anaemia". Doctors use free text fields to describe patient diagnosis in the EHRs making recognizing diseases from hundreds of texts to transform them to ICD a real challenge.

Accordingly, the diseases need to be annotated in the diagnosis texts, then they must be transformed to ICD automatically. The diagnosis texts dataset has been extracted from medical polyclinics EHRs in an excel file including 5000 texts. However, the diagnosis texts have some issues, such as missing values, medical abbreviations, grammar mistakes, and ambiguous values that doesn't refer to a specific disease term for example "pain". The diagnosis dataset has been processed manually to overcome these issues by eliminating errors, cleaning missing values, filtering medical abbreviations and other inaccurate data to get suitable 1000 records for tagging process. Table 3 shows a sample of the 1000 diagnosis text.

TABLE 3: A SAMPLE OF DIAGNOSIS DATASET.

| Diagnosis |
| --- |
| Acute lower dysphagia since yesterday evening after eating meat |
| The result of the analysis is Sickle-Cell Anaemia |
| Gastroenteritis |
| Stage 3 of Chronic Kidney Disease |
| Old known hypertension |
| Ischemic Heart Disease |
| New discovered hypertension + stroke |
| Chronic inactive hepatitis B |
| uncontrol Insulin-dependent diabetes mellitus |
| Follow up Diabetes mellitus type I /Primary hypothyroidism |

*B. Implementation Component*

This paper develops a web application that normalizes age, gender, and diagnosis dates. Then, diseases are annotated in the diagnosis texts to be transformed into ICD-10. The annotation process will be evaluated to measure performance accuracy. Finally, the web application allows downloading the standard demographics and diagnoses to be used as disease statistics in an interactive visualization platform.

Developing the web application is conducted using cloud computing server and Python and R operational environment. Amazon Web Services (AWS) has been customized to provide certain server instance for an appropriate operational environment. AWS provides multiple server instances with different features.

P2.xlarge is one of the available instances that has been allocated to be the web application server for this paper. P2.xlarge is ideally suited for machine learning, high-performance databases, perfect efficiency, and other server-side workloads requiring massive processing power. Jupyter Notebook is allocated to develop the web application on the P2.xlarge server. Jupyter Notebook used as an open-source web application to develop the system using live Python codes and machine learning tools which are shown in detail [27]:

1. Pre-processing: uses normalization regular expressions of R to produce a normalized data frame of the EHRs data. The data frame offers data type for each data element such as provide integer for all Age values and text for all Gender values. Since the system developed using Python, the R regular expressions are embedded into Python using rpy2 package. Rpy2 allows that any module installed for the R system can be used from within Python.

2. Annotation: uses SpaCy Python library for creating, training, and testing machine learning NER model to extract diseases from given diagnosis text of the EHRs

3. Transformation: uses ICD-10 API and NEL approach to match diseases with ICD-10 KB into transform them to standard form.

4. User interface: The user can use the system as a web application with an interface to be able to upload a CSV file.

As a result, the previous processes are performed on the data in the CSV file, which can be presented in the interactive dashboard using the Tableau application. Fig.2 shows the steps of developing the web application of the proposed model.

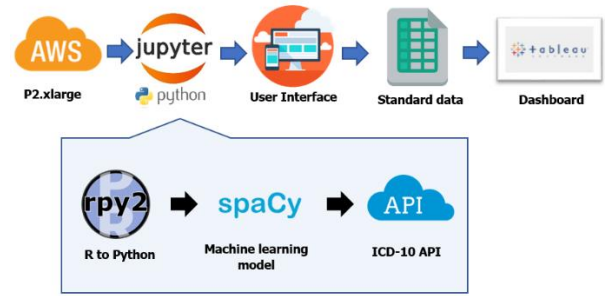

Fig. 2. Developing the web application of the proposed model.

## IV. SYSTEM IMPLEMENTATION

This section describes the system implementation of each phase and its processes to illustrate web application development steps with input/output examples.

### A. Pre-processing:

As shown in Fig.3, Pre-processing is applied to remove missing values from the medical dataset that include four attributes (Gender, Age, Diagnosis Date, and Diagnosis). Then, select Gender, Age, and Diagnosis Date attribute to normalize their values with specific normalization expressions for each.

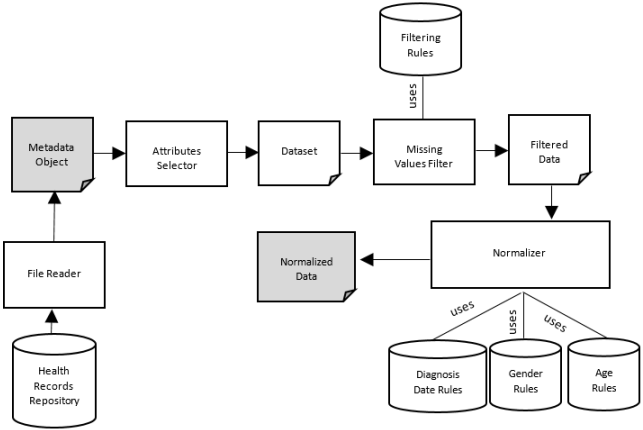

Fig.3. Pre-processing module.

Fig.4 and Fig.5 show a sample of input and output of pre-processing including 300 records of the EHRs medical dataset.

| Gender | Age | Diagnosis | Diagnosis Date |
|---|---|---|---|
| F | 21 | The patient's condition results in Diabetes Me... | 10 years ago |
| F | 56 | Tonsilitis | 8/4/1439 |
| F | 56 | Coryza | 8/4/1439 |
| F | 18 | Cystitis | 8/4/1439 |
| F | 23 | migraine | 8/4/1439 |
| f | 19 years | Arthitis | 6/4/1439 |
| f | 42 years | Siuusitis | 6/4/1439 |
| f | 34 years | Upper Respiratory Tract | 6/4/1439 |
| F | 20 years | Upper Respiratory Tract | 6/4/1439 |
| f | 26 years | Anxiety | 8/4/1439 |
| f | 26 years | The disease is Gastroenteritis | 8/4/1439 |
| f | 39 years | Siuusitis | 9/4/1439 |
| M | 22 | Hypertension | 14/5/1439 |
| F | 52 | The disease is Diabetes Mellitus 2 | 6/4/1439 |
| F | 52 | osteoarthritis | 6/4/1439 |
| F | 40 | Hyperthyroidism | 6/4/1439 |

Fig.4. Pre-processing input.

| Gender | Age | Diagnosis | Diagnosis Date |
|---|---|---|---|
| Female | 56 | Tonsilitis | 08/04/1439 |
| Female | 56 | Coryza | 08/04/1439 |
| Female | 18 | Cystitis | 08/04/1439 |
| Female | 23 | migraine | 08/04/1439 |
| Female | 19 | Arthitis | 06/04/1439 |
| Female | 42 | Siuusitis | 06/04/1439 |
| Female | 34 | Upper Respiratory Tract | 06/04/1439 |
| Female | 20 | Upper Respiratory Tract | 06/04/1439 |
| Female | 26 | Anxiety | 08/04/1439 |
| Female | 26 | The disease is Gastroenteritis | 08/04/1439 |
| Female | 39 | Siuusitis | 09/04/1439 |
| Male | 22 | Hypertension | 14/05/1439 |
| Female | 52 | The disease is Diabetes Mellitus 2 | 06/04/1439 |
| Female | 52 | osteoarthritis | 06/04/1439 |
| Female | 40 | Hyperthyroidism | 06/04/1439 |

Fig.5. Pre-processing output.

#### 1. Selecting attributes:

The medical dataset will be imported as CSV file to the system that checks four attributes to be selected including Age, Gender, Diagnosis Date, and Diagnosis. The four attributes may include some missing values which need to be cleaned.

#### 2. Removing missing values:

Removing missing values will filter all the four attributes to detect any missing value to remove it with its entire row. That because each attribute missing value in one row will affect the other attributes completeness as they associated with one patient. All patient information should be complete, including demographics and diagnosis. Filtering missing values may reduce the medical dataset rows because those have missing values cannot be replaced, but it does not affect the number of attributes.

#### 3. Normalizing demographics values:

Normalization is used to scale Gender, Age, and Diagnosis Date values, so they fit in a specific range. Adjusting the value range is very important for the attributes that contain different units and scales. Normalizing the three attributes is applied through rue-based approach using R regular expressions depending on.

- Gender:

The problem is about expressing gender values (Female/Male) with multiple expressions (i.e. "F" or "female" refer to "Female"). Therefore, specific gender formats that commonly found in the EHRs to represent gender data have been allocated for the normalization process. There are four common female formats (F, f, female, and Female), and four male formats (M, m, male, Male). By selecting Gender attribute, each value in the attribute will be matched with female and male formats to determine its category. Then it will be replaced with either "Female" or "Male" format using a regular expression. Table 4 shows Gender values before and after normalization.

TABLE 4: GENDER NORMALIZATION.

| Not normalized values | Normalized values |
|---|---|
| m | Male |
| f | Female |
| F | Female |

- Age:

Age normalization is considered more complicated since age values include four patterns in the EHRs. The four age patterns are integers, fractions, years or months, words and letters. For example, age can be expressed as "26", "20 m", "18 yrs", and " 1 1/2". Age normalization depends on determining the age pattern then replace it with standard integer format using a rule-based regular expression. Table 5 shows each age pattern with its normalization rule.

TABLE 5: AGE NORMALIZATION.

| Age pattern | Pattern value | Normalized values | Normalization rule |
|---|---|---|---|
| Integer | [16] | 16 | Remain as integer value. |
| Years | [16 y] or [16 years] | 16 | Remove words or letters that refer to years. |
| Months | [4 m] or [4 months] | NA | Replaced with "NA" as a missing value. |
| Fractions | [2 1/2] | 2 | Remove the fraction to get integer only. |

However, age values cannot be replaced by taking the average due to the importance of having real values. For example, Diabetes can be found in young people and adults, so the real values representing patients' ages must be considered. Therefore, all values for ages 1 year and older will be considered, while other values for babies who have their ages in months, which is less than one year, will be filtered out.

- Diagnosis date:

The diagnosis date needs to be normalized via rule-based regular expression in two steps. First, check the diagnosis date value to see if it is represented in date form or text form. If it represented in date form, only the "/" character will be acceptable for date structure values, so another character such as "-" will be replaced with the "/". Second, if it represented in text form such as " 5 years ago", it will be filtered out as the missing value. Table 6 shows diagnosis date values before and after normalization.

TABLE 6: DIAGNOSIS DATE NORMALIZATION.

| Not normalized values | Normalized values |
|---|---|
| 08-7-1439 | 8/7/1439 |
| 11/8/1439 | 11/8/1439 |
| more than 10 years | NA |

Considering the completeness of the dataset, each "NA" missing value in the Age, Gender, and Diagnosis date will be cleaned by filtering out using conditional expression.

*B. Annotation:*

As there is no specific value for each diagnosis text to be replaced with demographics values, diagnosis texts need the annotation process to extract disease names from them. Annotation process depends on selecting Diagnosis attribute first, then apply machine learning recognition model to recognize diseases names from each diagnosis text. Generating the machine learning recognition model requires a diagnosis dataset of annotated diagnosis texts.

As there is an available diagnosis dataset including 1000 diagnosis texts, annotating 1000 texts manually is a complex task as well as time-consuming due to the dataset size. Each disease name in each diagnosis text needs to be annotated by distinguishing it with a "Disease" tag. A machine learning annotation application called "Datatruks" has been used, which provides different annotation services such as Video Classification, Text summarization, or Named Entity Recognition (NER) tagging and more [28]. This paper applied NER tagging to tag all diseases names in the 1000 diagnosis texts automatically.

NER (also known as entity identification, entity chunking and entity extraction) is an information extraction task that responsible for locating and classifying named entities in text into pre-defined categories such as diseases. NER systems use linguistic grammar-based techniques and statistical models such as machine learning. Statistical NER systems usually require a huge amount of manually annotated training data. Thus, some tools have been suggested to avoid part of the annotation effort, such as the Dataruks annotation tool. [29]. NER training process in Datatruks will generate training dataset that includes tagged disease in the diagnosis texts through the steps shown in Fig.6:

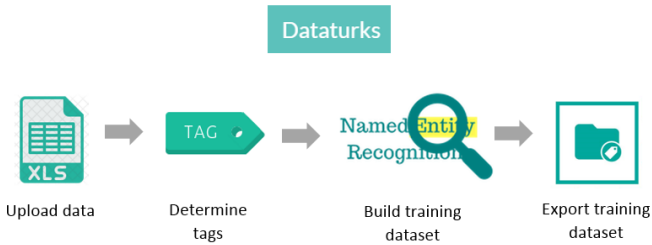

Fig.6. NER tagging steps.

The first step is uploading the diagnosis texts as an Excel file to the Datatruks. Then, determining required tags, the diagnosis dataset needs one tag to annotate diseases which is "Disease". After that, applying NER tagging in each text to tag any disease name to generate a training dataset. The tagging step is carried out under medical consultant supervision of a specialist doctor to ensure that the disease names are correctly identified in the texts. Finally, the training dataset will be exported in JSON format to be embedded in the system.

Since the system is based on Python, JSON doesn't support Python tuples. Although Dataturks NER JSON output is very close to the format used by SpaCy, SpaCy uses Python tuples; hence, Dataturks JSON training dataset needs to be converted to SpaCy training dataset using Python function [30]. After getting the SpaCy training dataset, there is a need for a coding scheme for encoding entity annotations as token tags. The begin In Last Unit Out (BILUO) scheme is used by NER SpaCy to explicitly mark boundary tokens. The BILUO scheme is considered easier than other schemes, such as the Begin, In, Out (IOB) scheme, which is more complex and difficult to learn. NER SpaCy has divided the dataset into training data (700 records) and testing data (300 records) to develop and test the Spacy machine learning model. Table 7 shows two examples of training datasets with "ANXIETY" diseases in JSON and SpaCy formats.

TABLE 7: JSON AND SPACY TRAINING DATASET.

| | |
|---|---|
| JSON | `{"content": "ANXIETY","annotation":[{"label":["Disease Name"],"points":[{"start":0,"end":6,"text":"ANXIETY"}]}],"extras":null,"metadata":{"first_done_at":1540994912000,"last_updated_at":1540994912000,"sec_taken":81,"last_updated_by":"m43sesOApAVsTFJigk3pvGtLSOV2","status":"done","evaluation":"NONE"}}` |
| SpaCy | `[('ANXIETY', {'entities': [(0, 7, 'Disease Name')]}),` |

Fig.7 shows how the system will recognize diseases form Diagnosis attribute texts using SpaCy machine learning model.

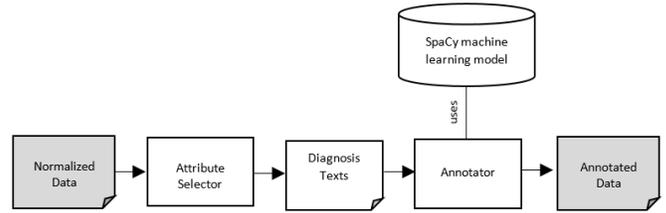

Fig.7. Annotation module.

1. *Select Diagnosis attribute:*

Since annotation is responsible for recognizing diseases from free text, the Diagnosis attribute will be selected as a target, ignoring Gender, Age, and Diagnosis date attributes.

2. *Annotator:*

Annotator uses machine learning Spacy model to recognize each disease name in a given text in the Diagnosis attribute that includes 241 texts out of 300 because of removing missing values. For the text that provides for more than one disease, the annotator will extract each disease as an output from that text. Table 8 shows some examples of given diagnosis text and diseases names that have been recognized.

TABLE 8: DIAGNOSIS TEXT' ANNOTATION.

| Diagnosis text | Diseases Names |
|---|---|
| Referred from shortness of breath/ pulmonary embolism /accepted by medical. | shortness of breath |
| | pulmonary embolism |
| phobic anxiety with major depressive disorder. | phobic anxiety |
| | major depressive disorder |
| Colon cancer for liver evaluation | Colon cancer |

*C. Transformation:*

As shown in Fig.8, the transformation is conducted through the following steps: generate two new attributes of ICD-10 Name and ICD-10 Code. Then, look-up each recognized disease name with ICD-10 Knowledge Base using the NEL approach. After that, taking the most priority result of linking process and assigning the ICD-10 name and code as values of the ICD-10 Name and Code attribute.

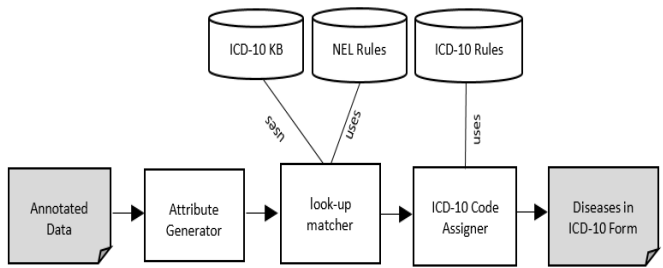

Fig.8. Transformation module.

*1. Attribute generator:*

So, three attributes—ICD-10 Name, ICD-10 Code, and ICD-10 Category — have been generated to be filled with their values after the transformation process. Fig.9 shows the output of the transformation model, which includes missing values due to the absence of a result matching the recognized disease name with ICD-10 KB.

| Gender | Age | Diagnosis | Diagnosis Date | ICD_10 Code | ICD_10 Name | ICD_10 Category |
|---|---|---|---|---|---|---|
| Female | 56 | Tonsilitis | 08/04/1439 | | | |
| Female | 56 | Coryza | 08/04/1439 | | | |
| Female | 18 | Cystitis | 08/04/1439 | A06.81 | Amebic cystitis | A06 |
| Female | 23 | migraine | 08/04/1439 | G43.B1 | Ophthalmoplegic migraine, intractable | G43 |
| Female | 19 | Arthritis | 06/04/1439 | | | |
| Female | 42 | Siuusitis | 06/04/1439 | | | |
| Female | 34 | Upper Respiratory Tract | 06/04/1439 | J39.9 | Disease of upper respiratory tract, unspecified | J39 |
| Female | 20 | Upper Respiratory Tract | 06/04/1439 | J39.9 | Disease of upper respiratory tract, unspecified | J39 |
| Female | 26 | Anxiety | 08/04/1439 | F41.1 | Generalized anxiety disorder | F41 |
| Female | 26 | The disease is Gastroenteritis | 08/04/1439 | | | |
| Female | 39 | Siuusitis | 09/04/1439 | | | |
| Male | 22 | Hypertension | 14/05/1439 | I15.0 | Renovascular hypertension | I15 |
| Female | 52 | The disease is Diabetes Mellitus 2 | 06/04/1439 | | | |
| Female | 52 | osteoarthritis | 06/04/1439 | M15.4 | Erosive (osteo)arthritis | M15 |
| Female | 40 | Hyperthyroidism | 06/04/1439 | P72.1 | Transitory neonatal hyperthyroidism | P72 |

Fig.9. Transformation output.

*2. Disease name look-up matcher:*

The look-up process matches each recognized disease name with ICD-10 KB using the NEL technique. ICD-10 KB contains a huge diseases classification, including disease information and standard code following the name. The look-up matcher will get the recognized disease name and use it as a search term to check whether it is matched with any term in the ICD-10 KB or not depending on the most priority ranking. The priority ranking is an automatic procedure in ICD-10 API that can sort the matching result depending on the matched vocabulary between the research term and KB terms. Table 9 Shows an example of look-up matching result of " diabetes mellitus type 1".

TABLE 9: Look-up matching result.

| Search term (recognized disease) | Look-up matching results |
|---|---|
| Diabetes mellitus type 1 | 1. E10.9 Type 1 diabetes mellitus without complications<br>2. E10.21 Type 1diabetes mellitus with diabetic nephropathy<br>3. E10.36 Type 1 diabetes mellitus with diabetic cataract<br>4. E10.41 Type 1 diabetes mellitus with diabetic mononeuropathy |

The result of the search term " diabetes mellitus type 1" will be a set of ranked ICD-10 terminology, depending on the highest priority. Assigning one of the highest priority results for the disease name needs to follow some rules and conditions.

*3. Assigning ICD-10 name, code, and category:*

A set of conditional rules has been allocated to assign the most priority ICD-10 name and code. The conditional rules include selecting the first ranked result as a matching result, getting only the ICD-10 code and name, and ignoring the rest of the disease details and information. After that, take a copy of the found ICD-10 code and split it by removing the numbers after the comma to get the basic disease category without branching. In case there are two search terms, the assigner should get the first-ranked result of the first term and then the first-ranked result of the second term.

The assigner will store the ICD-10 names and codes in the ICD-10 Name and ICD-10 Code attribute and store the split ICD-10 code in the ICD-10 Category attribute for each search term. But, if there is no result for matching, ICD-10 name and code will be "NA" as a missing value. The missing values in ICD-10 Name and Code attributes will not be cleaned, the reason behind that is to leave an option for manual ICD coding and to maintain the reliable number of diseases in the

The system's output is a standard data frame of 7 attributes, including Age, Gender, Diagnosis date, Diagnosis, ICD-10 Name, ICD-10 Code, and ICD-10 Category. The standard data frame can be downloaded as a CSV file and presented as statistics in a visualization application. Table 10 shows a sample of the standard data frame.

TABLE 10: STANDARD DATA FRAME.

| Age | Gender | Diagnosis Date | ICD-10 Code | ICD-10 Name | ICD-10 Cat. |
|---|---|---|---|---|---|
| 59 | Female | 08/04/1439 | Q30.0 | Choanal atresia | Q30 |
| 48 | Female | 14/04/1439 | Y95 | Nosocomial condition | Y95 |
| 63 | Female | 13/04/1439 | P70.2 | Neonatal diabetes mellitus | P70 |
| 60 | Male | 13/05/1439 | Q30.0 | Choanal atresia | Q30 |
| 60 | Male | 12/05/1439 | - | - | - |
| 55 | Female | 08/05/1439 | P72.1 | Transitory neonatal hyperthyroidism | |
| 62 | Female | 08/05/1439 | - | - | - |

*D. Evaluation:*

This paper seeks to evaluate two types of diseases recognition systems which are dictionary-based systems and machine learning systems. The evaluation process will compare the SpaCy machine learning model that was generated for the annotation process with MetaMap. MeataMap is a dictionary-based system that will be compared with the SpaCy system in terms of performance accuracy through some steps. First, use the same diagnosis texts for both systems, then monitor the performance measures manually and compare the accuracy results as shown in Fig.10.

*1. Diagnosis texts:*

As mentioned in the annotation phase, the Diagnosis attribute values will be used for diseases recognition by SpaCy system. Diagnosis attribute includes 242 diagnosis texts that embedded to Spacy system automatically. SpaCy performance results will be recorded in a specific table for comparison process. However, since MeatMap is available for manual entry of single block of text, 242 texts will be entered to MetaMap manually one by one.

*2. Monitor performance results:*

Each diseases recognition result in both systems will be sorted as true or false result manually. The true results involve any recognized disease, either exactly or partially matching the disease in the diagnosis text, provided that the result indicates the same range of disease. For example, if the result of recognizing diseases from " Patient has Diabetes Mellitus 2" is " Diabetes Mellitus", it will be considered as a true result due to referring to the same disease type. False results include any recognized diseases that partially matched the original disease to refer to vague terms. For example, if the result of recognizing diseases from " Gastroesophaged Reflux Disease" is "Disease", it will be considered as a false result. In addition, false results include any recognition null result or any word that does not belong to the disease and has been extracted as a disease. Table 11 shows the performance results of SpaCy and MetaMap.

TABLE 11
SPACY AND METAMAP PERFORMANCE RESULTS.

| Age | Gender | Diagnosis Date | ICD-10 Code |
|---|---|---|---|
| 59 | Female | 08/04/1439 | Q30.0 |
| 48 | Female | 14/04/1439 | Y95 |
| 63 | Female | 13/04/1439 | P70.2 |
| 60 | Male | 13/05/1439 | Q30.0 |
| 60 | Male | 12/05/1439 | - |
| 55 | Female | 08/05/1439 | P72.1 |
| 62 | Female | 08/05/1439 | - |

*3. Evaluation results:*

The true and false result values for both systems are collected to estimate their accuracy. The accuracy measure is estimated by dividing the sum of true results on the number of the given texts (242 texts) for both systems individually.

*E. Visualization:*

The standard data frame will be exported from the web application as a CSV file and imported to interactive visualization platform to be displayed as statistics as shown in Fig.11.

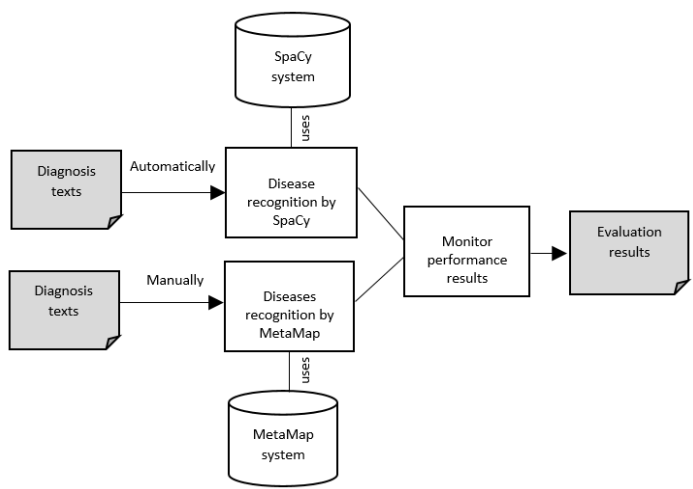

Fig.10. Evaluation module.

Fig.11 presents the results of sentiment analysis using Pie chart. It shows the percentage of positive, negative and neutral views as 55%, 44%, 1% respectively.

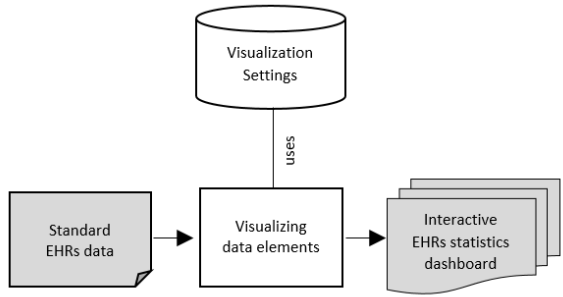

Fig.11. Visualization module.

*1. Standard EHRs data:*

The web application provides a download option for the standard data frame as a CSV file, which includes 7 attributes: age, Gender, Diagnosis date, Diagnosis, ICD-10 Name, ICD-10 Code, and ICD-10 Category. The user can utilize the standard data using a visualization application to present the data as useful statistics of ICD-10 diseases with their demographics.

*2. Visualizing data elements:*

We used Tableau as a visualization platform to present the standard EHRs data as statistics. Tableau software produces interactive data visualization dashboards of specific dimensions such as diseases and age categories. The EHRs data elements will be visualized as dimensions using visualization settings. For example, age values will be divided into specific categories such as 30-49, 50-69, and so on. The dashboard will be designed to present all the 7 attributes elements in an interactive way.

*3. Interactive EHRs statistics dashboard:*

The result is an interactive dashboard of the EHR statistics, where interactive means the availability of querying about diseases or determining specific age categories to be presented. The dashboard interacts with each query and changes the presented elements according to that. The interactive visualization of the standard EHRs data will support research activities by providing more reliable diseases data. Researchers can analyse the statistical results and monitor the gender or age groups most affected by a disease. They can also measure the prevalence of a specific disease during each month and the possibility of sharing and comparing the findings with other medical or research bodies as they are consistent and standardized in ICD-10 form. Fig.12 shows a default design of standard disease statistics in interactive dashboards.

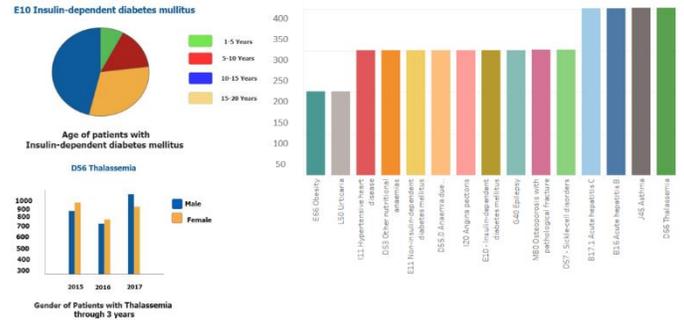

Fig.12. Disease statistics in interactive dashboards.

## V. EVALUATION AND RESULTS:

This section outlines the evaluation and results of the proposed model implementation, alongside showcasing the standard data through an interactive visualization platform.

*1. Evaluation and results:*

Developing the web application user interface is divided into two sections: interface design and interface operations. The interface design includes two pages. The first page presents an icon to select the data file from the computer. The second page presents the result of data processing as a standard data frame with a download icon to download it on the computer. Fig.14 and Fig.15 show the two pages of the web application interfaces.

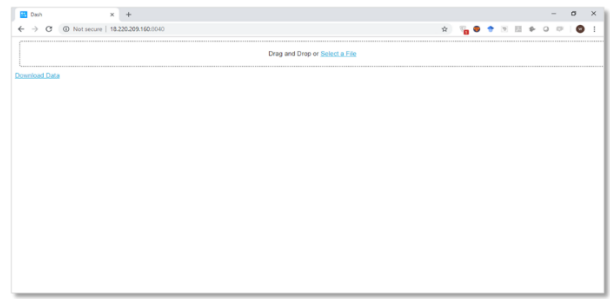

Fig.14. First interface.

Fig.15 Second interface.

The annotation performance is evaluated by comparing the diseases recognition results of a machine learning system (SpaCy annotator) and a dictionary-based system (MetaMap annotator). The evaluation depended on annotating 241 diagnosis texts in both systems and monitoring the results of true or false results to measure the accuracy. Where the diagnosis texts are annotated by SpaCy annotator automatically using NER rules and SpaCy training dataset in the annotation model. However, for MeatMap, the diagnosis texts are manually annotated text by text in the MetaMap demo. Fig.16 and Fig.17 show a sample of diseases recognition results of SpaCy and MetaMap annotators.

Fig.16. SpaCy annotator result.

Fig.17. MetaMap annotator result.

Correct boundaries of the recognized diseases are considered as the exactly matched result, and the ones having boundary overlap with the correct annotations but not exactly matched are considered as the partially matched result. The accuracy is measured individually for both systems by dividing the total of the true result (the recognized diseases) including the exactly and partially matched results on the total of the diagnosis texts:

**Accuracy = Total true results / Total results**

Table 12 presents the accuracy measurements of SpaCy and MetaMap annotators regarding the 241 diagnosis texts, which contain 241 disease terms.

TABLE 12: Accuracy measurements.

| Annotator | True result | False result | Accuracy |
|---|---|---|---|
| SpaCy (Machine learning annotator) | 197 | 44 | 81 % |
| MetaMap (Dictionary-based annotator) | 162 | 79 | 67 % |

The MetaMap annotator can correctly annotate 159 exactly matched ones and 19 partially matched ones. The SpaCy annotator can correctly annotate 183 exactly matched ones and 57 partially matched. The accuracy of the SpaCy annotator is 81 % outperformed than the MetaMap which is 67 %.

2. *Visualization:*

The web application provides a standard EHRs data which include patients' diagnosis in ICD-10 form and standard demographics. The standard EHRs data can support research activities by providing diseases and population interactive statistics. We used Tableau as an interactive visualization platform to represent it and offer different interactive dashboards with multiple choices of data dimensions. After downloading the standard EHRs data file from the web application, it is imported in Tableau to display the data frame and design the dashboard. The dashboards display the statistics with checkboxes or lists to determine specific items, such as selecting a disease or gender type, so the dashboard will interact with that by displaying the updated results of statistics.

There are multiple dashboards that have designed such as EHRs diseases statistical dashboard. As shown in Fig.18 the dashboard displays comprehensive statistics including diseases with ICD-10 names, ICD-10 categories, age range, gender, and diagnosis date as well as the EHRs data frame itself.

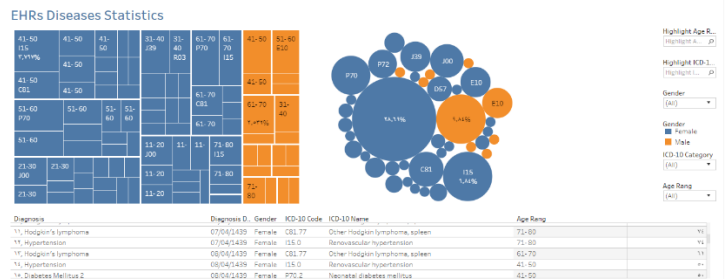

Fig.18. Comprehensive statistics.

## V. CONCLUSION

In general, the EHRs data are unsuitable for secondary use due to the use of natural language by doctors, making them in various formats and free text. However, the EHRs data include two main components that are commonly captured for research, which are diagnosis data and demographics. For diagnosis, the paper has developed SpaCy (machine learning model) to recognize diseases from the diagnosis text as they are written in free text fields in the EHRs. The SpaCy model is compared with the (dictionary-based model) MetaMap to evaluate their accuracy. The issue found with the SpaCy model's false results involved considering unrelated words as a part of the recognized disease name. For example, recognize " has diabetes mellitus" as a disease name from this text: " the patient has diabetes mellitus", so "has" is considered as a part of that disease name. The reason behind that is that the SpaCy model depends on the position of disease words in the text and keeps training on that. Thus, SpaCy model may recognize the disease name (specific words) in some sentences, but it doesn't recognize the same disease name in others. This means that each text has different words surrounding the disease name, which confuses the model in determining the disease boundaries. This issue is caused by using informal writing of diagnosis descriptions by doctors.

The issue found with MetaMap model false results is giving null values referring to an unfound disease name in the text, although the text includes a disease name. That is because MetaMap depends on matching the disease name in the text with specific terms in a medical dictionary to be recognized. Therefore, MetaMap couldn't recognize some diseases at all due to the limitation of particular dictionary terms. Furthermore, MetaMap has developed to recognize diseases and other biomedical concepts, such as biological elements that cause dispersion to the system. So, some false results of disease recognition in MetaMap are caused by splitting the disease name, which may include some words associated with biologically element, into two recognized concepts as "disease" and "biologically element". For example, if the disease name is "Iron anemia", it will be recognized as "Iron" and "anemia" separately. Where "Iron" refers to the biological element and "anemia" refers to disease. In contrast, SpaCy will recognize the full name of the disease even if it contains vocabularies for some biological element. In the end, the accuracy of SpaCy model can be improved by increasing the number of training dataset to train the model better. MeatMap is offered to recognize diseases from texts written in the official form, and this is not often available in the natural language of doctors.

Regarding the paper scope, there are some limitations shown below, following with recommendations:

1. The medical dataset gathered from the EHRs of polyclinics manually because it was not possible to export the disease data directly from the system. Most data consist of the patients of polyclinics. Therefore, the testing dataset is available in limited formats. So, there is a need to gather the medical dataset from more EHRs sources to discover additional variations of data formats and to improve the system accuracy.

2. The system depends on the attributes labeling in the medical dataset to be processed. For example, if the medical dataset file contains an attribute called "Sex" instead of "Gender", the system will not be able to accept the file. To overcome this, the system needs to use a Topic Deduction approach that can recognize the attribute by its values and not by its label itself.

3. Age normalization is limited in standardizing formats of age from one year to more and does not include age formats for children under one year. Age normalization can be improved by including more age formats such as months and days.

## VI. FUTURE WORK

For future work, there are three parts can be enhanced in this research seeking better work. First, more complex diagnosis texts should be included rather than short phrases. Such as using medical documents or long clinical notes to produce a more advanced machine learning model for diseases recognition. Second, providing an integrated web application that processes the EHR data and then displays them in interactive visualization platforms. This will make it easier for researchers to use a unified platform that provides standard EHRs data and display them as reliable statistics at the same time. Finally, the continuous evolution of ICD should be taken into consideration to support research activities and statistics with the latest international classifications of diseases.

## ACKNOWLEDGMENT

The authors would like to express their deepest gratitude to Dr. Majed Aadi Alshamari (Information Systems department at CCSIT), Dr. Ahlam Haddad, Dr. Afrah Almaleh, Dr. Misfer Mutlaq Aldossari (the Poly Clinics at King Faisal University), and to Dr. Lulwah Al Turki (King Abdulaziz Medical City), and to Dr. Waleed Alabbas (Data Scientist) in Saudi Arabia for supporting.